\documentclass{article}
\usepackage{spconf,amsmath,graphicx}
\usepackage{tabularx}
\usepackage{color,soul}

\usepackage[math]{cellspace}
\title{NeRD: a \textbf{Ne}ural \textbf{R}esponse \textbf{D}ivergence Approach to Visual salience detection}
\name{M.~J. Shafiee{$^\dagger$}, P. Siva{$^\ddagger$}, C. Scharfenberger{$^\dagger$},  P. Fieguth{$^\dagger$} and A. Wong{$^\dagger$}  \thanks{This work was supported by the Natural Sciences and Engineering Research Council of Canada, Canada Research Chairs Program, and the Ontario Ministry of Research and Innovation. The authors also thank Nvidia for the GPU hardware used in this study through the Nvidia Hardware Grant Program.}}
\address{{$^\dagger$} Vision \& Image Processing Research Group, Univeristy of Waterloo, Waterloo, Canada\\ {$^\ddagger$} Aimetis Corp.,
       Waterloo, Canada}
\begin{document}
%
\maketitle
\begin{abstract}
In this paper, a novel approach to visual salience detection via  \textbf{Ne}ural \textbf{R}esponse \textbf{D}ivergence (NeRD) is proposed, where synaptic portions of deep neural networks, previously trained for complex object recognition, are leveraged to compute low level cues that can be used to compute image region distinctiveness. Based on this concept , an efficient visual salience detection framework is proposed using deep convolutional StochasticNets.  Experimental results using CSSD and MSRA10k natural image datasets show that the proposed NeRD approach can achieve improved performance when compared to state-of-the-art image saliency approaches, while the attaining low computational complexity necessary for near-real-time computer vision applications.
\end{abstract}
%
\section{Introduction}
\label{sec:intro}
\vspace{-0.3 cm}
Salient object detection has seen significant attention in recent literature \cite{mai2014comparing,margolin2013makes,zhao2015saliency,wang2015deep}. Visual saliency detection, for salient object detection, aims to mimic the human visual system when identifying and segmenting the most salient objects in natural images or videos.

  Saliency detection/assessment can be formulated as  unsupervised~\cite{mai2014comparing,margolin2013makes,perazzi2012saliency,siva2013looking} or supervised \cite{HSD2013hierarchical,shi2013pisa,zhao2015saliency,wang2015deep,mai2013saliency}. Supervised algorithms attempt to train classifiers to directly detect salient objects,  modeling the background~\cite{mai2013saliency} or taking advantage of spatial constraints~\cite{shi2013pisa,yang2013saliency}. Recently, motivated by the success of deep learning~\cite{Bengio-2009}, the salience detection problem has been formulated as a learning task for a deep neural network to find the salient object in an image~\cite{zhao2015saliency,wang2015deep}.

However, saliency detection is mostly an unsupervised task in the human brain and a person does not explicitly learn how to identify salient object in an image. As a result the unsupervised approaches are of more interest due its consistency with nature. The most common approach here is to hand-craft image distinctiveness features inspired by the human visual system~\cite{TD2013statistical,LR2012unified,DRI2013salient}. However, these image feature are local features capable of identifying local salient points in images and are not suitable for identifying salient objects in images, which requires more global image level features. Then to obtain salient object, these hand-crafted local features are combined into global cues~\cite{HSD2013hierarchical,scharfenberger2015structure,cheng2015global}.

Hand crafted features identify local points of interest such as corners and edges, however they are crafted by our interpretation of the human visual system. We argue that these features are developed by the human visual system when people learn complex tasks such as object recognition. Later, these same low level features learned for object recognition are used to identify salient objects in images. As such, we present \textbf{Ne}ural \textbf{R}esponse \textbf{D}ivergence (NeRD) which utilizes the neural responses, learned for the object recognition task, as features for identifying salient objects in natural images.


\textbf{Ne}ural \textbf{R}esponse \textbf{D}ivergence (NeRD) is provided in an unsupervised structure-based saliency detection~\cite{scharfenberger2015structure} framework to assign higher saliency value to the regions belongs to salient object in the image. The synaptic weights of receptive fields in the proposed NeRD approach are provided by pre-trained convolutional neural networks which make  the proposed algorithm independent from any saliency specific training procedure.

Saliency algorithms are quite often used as a pre-processing step for other computer vision algorithms, as such they require low computational complexity. In order to reduce the computational complexity of the convolutional neural networks, we make use of StochasticNets~\cite{shafiee2015stochasticnet} which use random graph theory to obtain deep neural networks that are sparsely connected. Using this sparse connectivity, the NeRD framework can identify salient objects with lower computational complexity than conventional convolutional neural networks.


\vspace{-0.4 cm}
\section{Methodology}
\label{sec:Method}
\vspace{-0.2 cm}
The proposed approach to assessing visual salience is described in detail as follows. First, the underlying motivation and concept behind \textbf{Ne}ural \textbf{R}esponse \textbf{D}ivergence (NeRD) is explained. Second, an efficient visual salience detection framework using deep convolutional StochasticNets~\cite{shafiee2015stochasticnet} is presented in detail.
\vspace{-0.6 cm}
\subsection{\textbf{Ne}ural \textbf{R}esponse \textbf{D}ivergence}
\vspace{-0.25 cm}
The proposed approach to salient region detection in images takes inspiration from the way the human brain learns to assess salience based on visual stimuli.  Visual salience assessment is a major attentional mechanism of the human brain, as it is very important to be able to rapidly identify predator, prey, food, and other objects necessary for survival in visually complex, cluttered environments.  As such, visual salience assessment has evolved into an unconscious, reactionary mechanism which allows the brain to localize regions of interest spontaneously such that limited cognitive resources can be focused on complex object recognition in such regions of interest.  While the primitive part of visual salience in the human brain is purely reactionary and developed as part of the evolutionary process, the more experiential part of visual salience can be also be attained through training.  It is this memory-dependent, experiential aspect of visual salience in the human brain that is particularly interesting, as human beings are not trained in a direct, binary fashion to determine whether an object is salient or not.  On the contrary, the improved ability to detect contextually-relevant salient objects through visual stimuli over time is an indirect consequence of the human brain being trained for complex object recognition within a given environment.  For example, numerical symbols can become salient as a consequence of a child directly learning to recognize numbers.  In essence, this experiential part of visual salience in the human brain can be viewed as an unsupervised process that leverages knowledge acquired through supervised training in the complex object  domain.

As such, we are inspired to adopt this indirect visual salience learning phenomena of the human brain to develop a new unsupervised approach to visual salience detection by leveraging synaptic portions of deep neural networks that have been directly trained for complex object recognition.  More specifically, the proposed approach leverages the first synaptic layer of a deep neural network, which has been demonstrated to model primitive, low-level representations of complex data.  In the context of visual stimuli, the first synaptic layer of a deep neural network trained for complex object recognition shows strong neural response to low-level, highly discriminative structural and textural characteristics of visual stimuli, which makes it well-suited for discriminating between different objects in a scene, or in our case between salient objects and non-salient background clutter.  As such, based on the neural responses for the entire scene attained using this first synaptic layer of a deep neural network, the proposed method then quantifies the distinctiveness of particular neural responses relative to other neural responses in the scene to assess how well these responses stand out, which is indicative of how visually salient they are within the scene.

One of the main benefits of the proposed \textbf{Ne}ural \textbf{R}esponse \textbf{D}ivergence (NeRD) approach to assessing visual salience is that one can take significant advantage of the wealth of training data as well as rapid advances in deep neural networks in the realm of complex object recognition to greatly improve visual salience detection performance over time.  A further benefit  is that, by sharing cognitive resources with a deep neural network trained for complex object recognition, one can form a unified framework that can identify regions of interest as well as perform object recognition on these regions in an efficient manner.
\vspace{-0.56 cm}
\subsection{Visual Salience Detection Framework}
\vspace{-0.2 cm}
Based on the concept of NeRD, we now present an efficient visual salience detection framework  designed specifically for the low computational complexity necessary for near-real-time computer vision applications.    First, the first synaptic layer of a deep convolutional StochasticNet~\cite{shafiee2015stochasticnet} that was previously trained for complex object recognition is used to obtain neural responses $f_k$ at each pixel $k$ of the image.  Second, a sparse neural response model composed of a set of representative neural response atoms $t_i$ is constructed based on the pixel-level neural responses, and the pair-wise neural response distinctiveness is computed between each pair of atoms in the sparse model.  Finally, the visual salience of each pixel $\alpha_k$ is then computed based on the conditional expectation of pairwise neural response distinctiveness given the atom that the pixel belongs to.  Each of these steps is designed with computational efficiency in mind and are described in detail below.

\begin{figure*}
\begin{center}
\includegraphics[width = 12 cm]{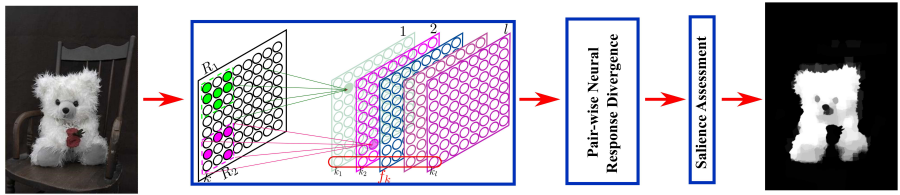}
\vspace{-0.4 cm}
\caption{ Salience assessment based on the NeRD framework. The neural response divergence is extracted based on the StochasticNet approach and the features are utilized to compute the saliency map for the input image. The neural response divergence feature $f_k$ has the same size as the number of receptive fields $\{R_1, R_2, ..., R_l\}$in the processing block.  }
\label{fig:NeRD}
\end{center}
\end{figure*}

\vspace{-0.6 cm}
\subsubsection{Neural Response Extraction via StochasticNets}

\vspace{-0.2 cm}
 One of the biggest obstacles to this neural response extraction process is the high computational complexity associated with popular deep neural networks such as deep convolutional networks~\cite{CNN2010convolutional} and deep belief networks~\cite{DeepbeliefNet2009convolutional}.  In order to attain low computational complexity, the proposed framework utilizes the concept of StochasticNets~\cite{shafiee2015stochasticnet,shafiee2015discovery}, where random graph theory is leveraged to stochastically form the neural connectivity of deep neural networks such that the resulting deep neural networks are highly sparsely connected yet maintain the modeling capabilities of traditional densely connected deep neural networks. Figure~\ref{fig:NeRD} demonstrates the flow-diagram of the  NeRD framework for extracting the neural response divergence and computing the image salience map.

In the proposed framework, a deep convolutional StochasticNet based the AlexNet~\cite{AlexNet2012imagenet} network architecture trained using the ImageNet dataset but with only 25\% neural connectivity is constructed, with the neural responses of its first processing block (i.e.,  the combination of convolutional layer, rectifier, normalization and the pooling layer) used for computing neural response divergence in the second step.  The utilization of such a sparsely-connected deep convolutional StochasticNet architecture fa leads to significantly reduced computational complexity, compared to state-of-the-art unsupervised visual salience methods.
\vspace{-0.4 cm}
\subsubsection{Pair-wise Neural Response Divergence using Sparse Neural Response Modeling}
\vspace{-0.1 cm}



The extracted neural responses for pixel $k$ encoded by $f_k$ is obtained as distinctive features to construct  the set of representative neural response atoms. The number of  distinctive features $|f_k| = l$ is  equal to the number of stochastic receptive fields $l$ in the processing block.
\begin{figure*}
\vspace{- 0.7 cm}
\begin{center}
 \setlength\tabcolsep{0.3 pt}
\begin{tabular}{cc}
\includegraphics[width = 6 cm]{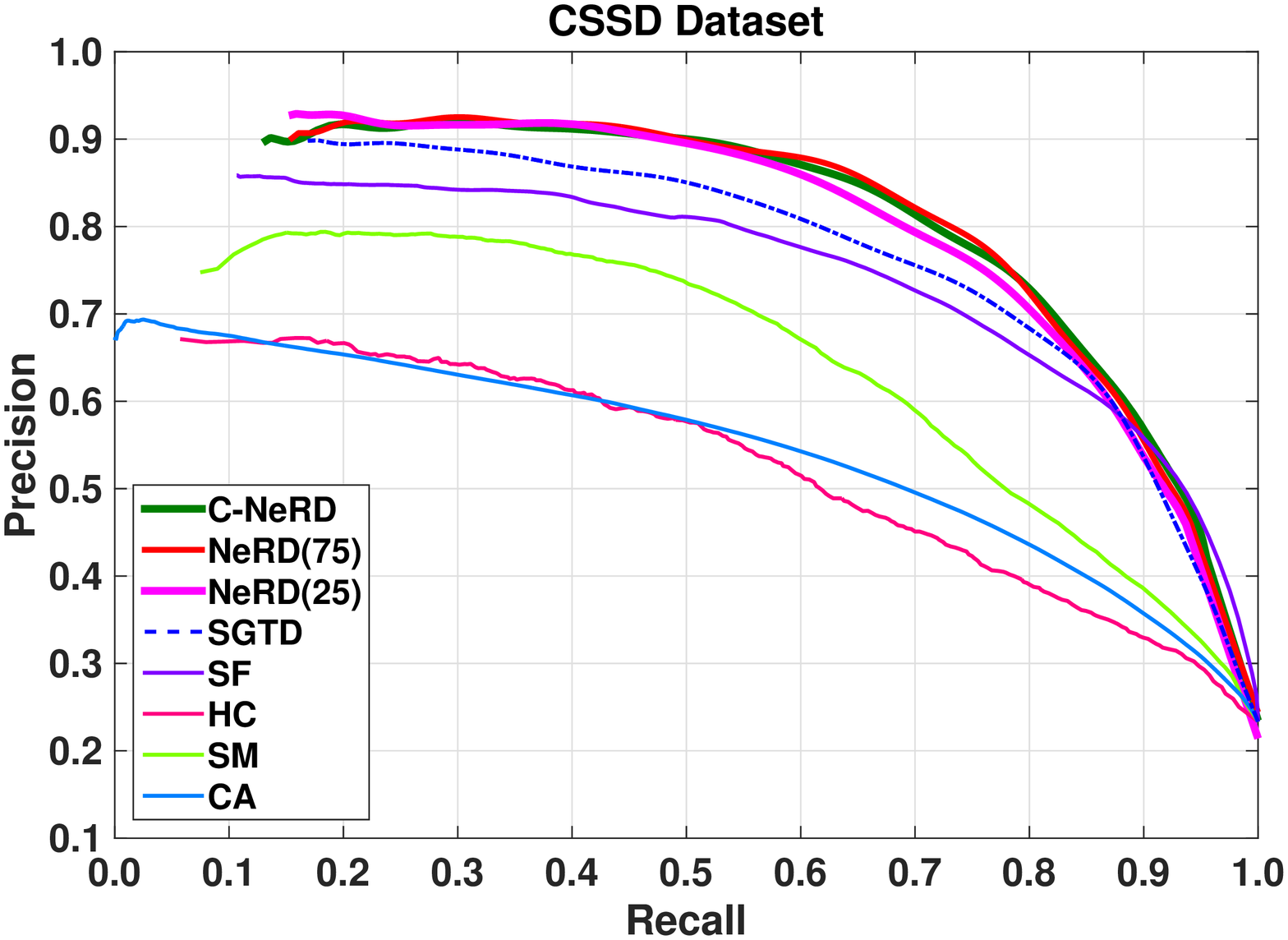}&
\includegraphics[width = 6 cm]{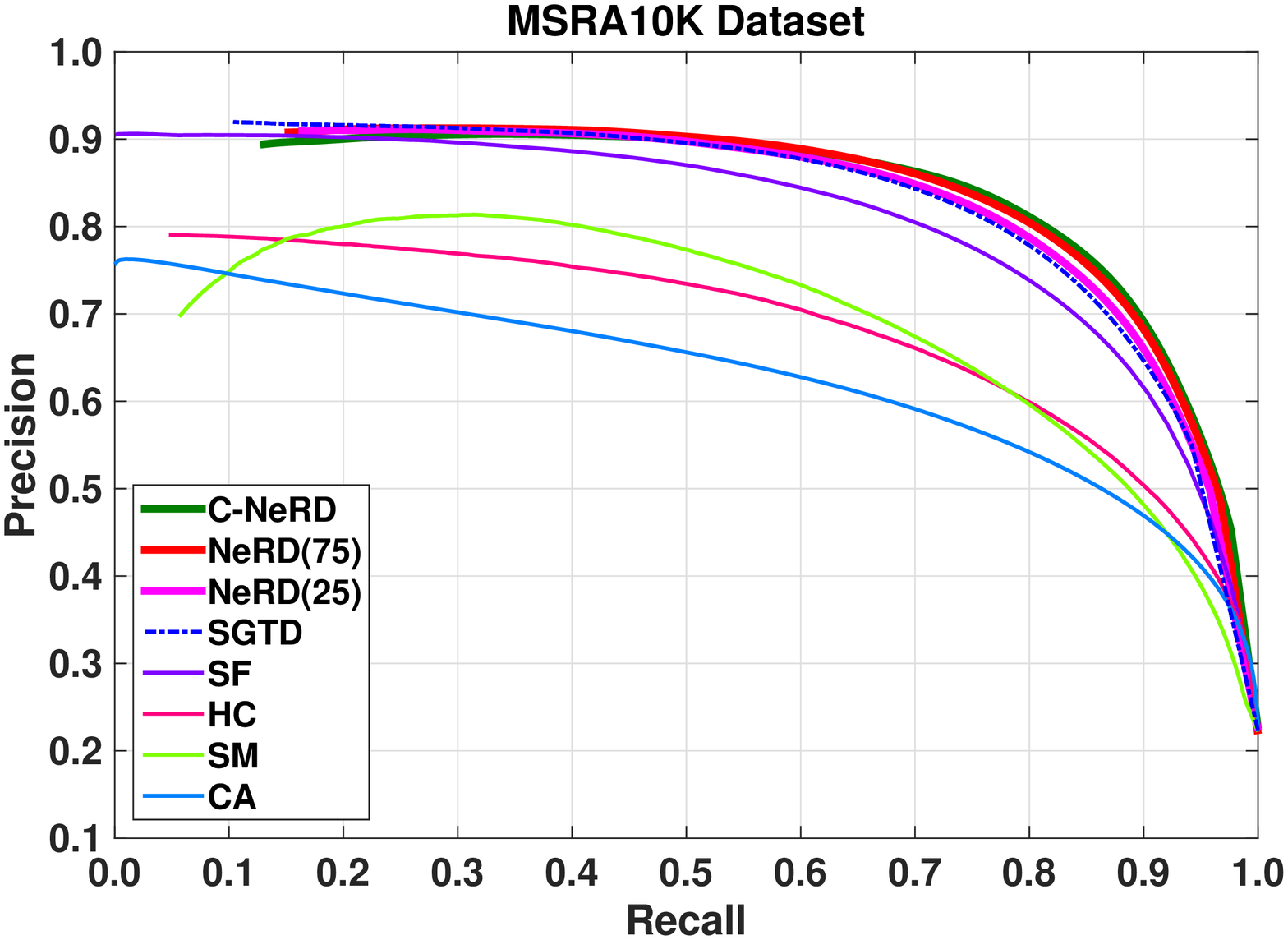}
\end{tabular}
\vspace{-0.5 cm}
\caption{ Precision and recall rates for all tested approaches based on  CSSD~\cite{HSD2013hierarchical}  and MSRA10k~\cite{MSRA2009frequency}. Our approaches (C-NeRD, NeRD(75) NeRD(25)) outperform  other state-of-the-art algorithms in both CSSD and MSRA10k datasets. }
\label{fig:PRCurve}
\end{center}
\vspace{-0.75 cm}
\end{figure*}

  To preserve the structural characteristic of the input image, the image is  decomposed into a set of image elements $\mathcal{E} = \{e_i\}_{i=1}^m$ where the elements $e_i$ and $e_j$ are decomposed in a way that local boundaries and fine structures are located in the ridge of two elements. The SLIC superpixel approach~\cite{achanta2012slic} is incorporated to find the set of elements $\mathcal{E}$ to preserved  the mentioned property.

  Given image $I_{w\times h}$ and generated distinctive features \mbox{$\mathcal{F} = \Big\{f_k\Big\}_{k=1}^{w \times h}$}, the atom neural response   $\mathcal{T}$ is constructed based on the elements $\mathcal{E}$. A simple element-wise averaging is utilized to compute the  atom neural response  $t_i \in \mathcal{T}$ corresponding to element $e_i$:
  \vspace{-0.3 cm}
  \begin{align}
  t_i = \frac{1}{z} \sum_{k \in e_i} f_k
  \end{align}
\noindent where  $z = |e_i|$ is a normalization,  the number of pixels belonging to element $e_i$.  Obtaining the  atom neural response  makes the computational complexity much cheaper since the number of  elements in $|\mathcal{T}| = m$ and \mbox{$m \ll w \times h$}.

   However to make the model more sparse, the  similar atom neural responses are combined and represented as a single atom neural response $\hat{t}_i $ by applying clustering. A k-means clustering algorithm is applied to  $\mathcal{T}$, generating  the sparse set of atom neural responses $\hat{\mathcal{T}}$,  where $|\hat{\mathcal{T}}| \ll |\mathcal{T}|$. Each element $\hat{t_i} \in \hat{\mathcal{T}}$ represents the distinctive features regarding a subset of \mbox{$\{e_i |e_i \in \mathcal{E}\}$}.  To this end the subset of $e_i$'s which are encoded by the sparse neural atom response $\hat{t_i}$ is represented \mbox{by $s_i$}.

\begin{figure*}
\vspace{- 0.99 cm}
 \centering
 \setlength\tabcolsep{0.1 cm}
{\renewcommand{\arraystretch}{0.3}
\begin{tabular}{cccccccccc}
\includegraphics[width = 1.1 cm]{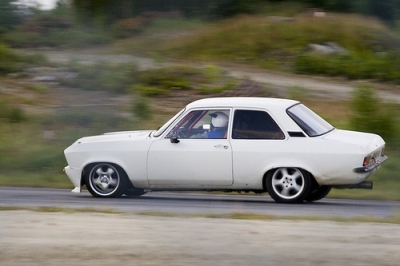}&
\includegraphics[width = 1.1 cm]{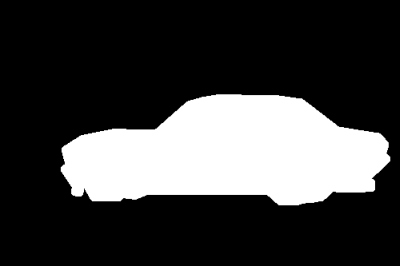}&
\includegraphics[width = 1.1 cm]{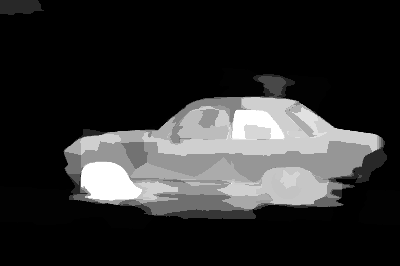}&
\includegraphics[width = 1.1 cm]{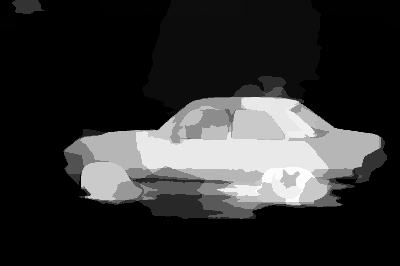}  &
\includegraphics[width = 1.1 cm]{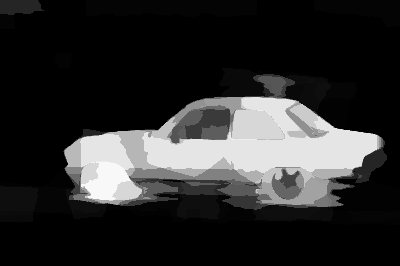}  &
\includegraphics[width = 1.1 cm]{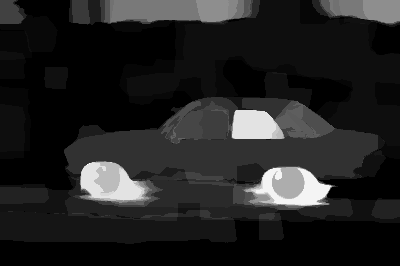}&
   \includegraphics[width = 1.1 cm]{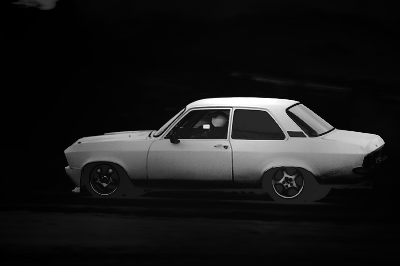}&
   \includegraphics[width = 1.1 cm]{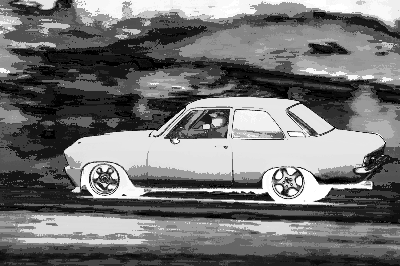} &
   \includegraphics[width = 1.1 cm]{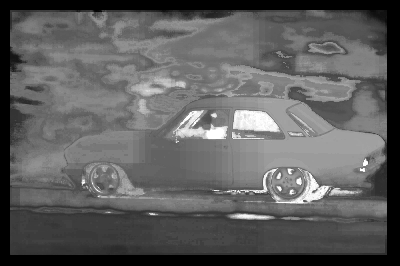} &
      \includegraphics[width = 1.1 cm]{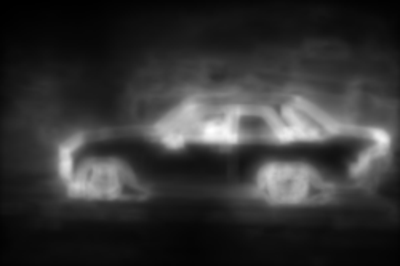} \\

\includegraphics[width = 1.1 cm]{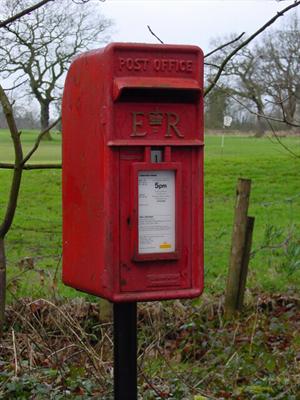}&
\includegraphics[width = 1.1 cm]{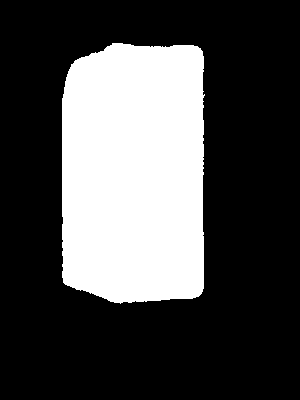}&
\includegraphics[width = 1.1 cm]{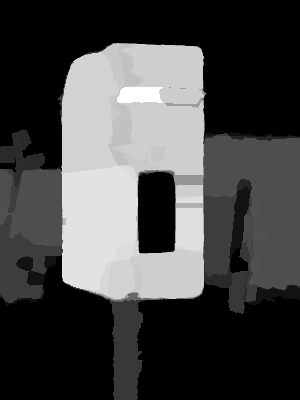}&
\includegraphics[width = 1.1 cm]{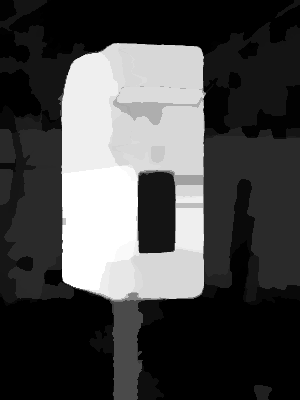}  &
\includegraphics[width = 1.1 cm]{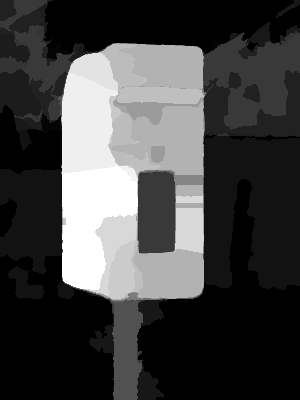}  &
\includegraphics[width = 1.1 cm]{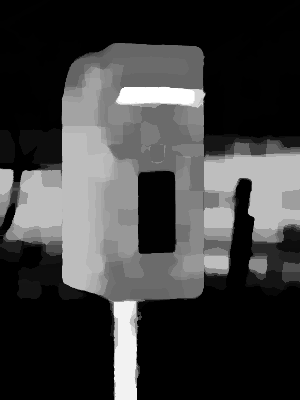}&
   \includegraphics[width = 1.1 cm]{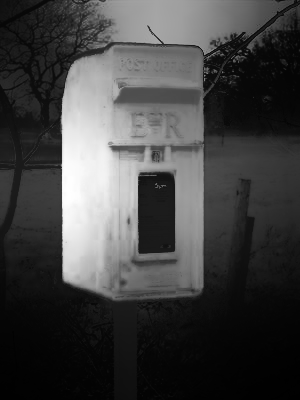}&
   \includegraphics[width = 1.1 cm]{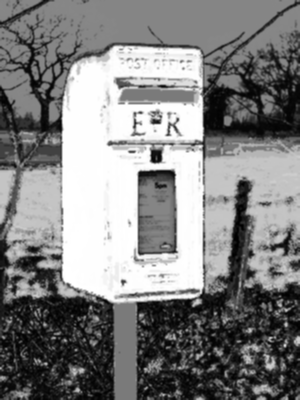} &
   \includegraphics[width = 1.1 cm]{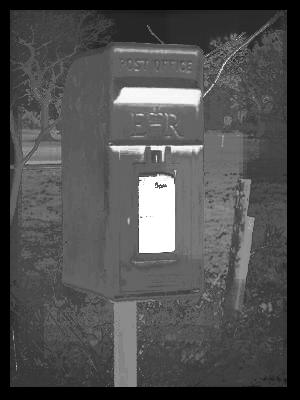} &
      \includegraphics[width = 1.1 cm]{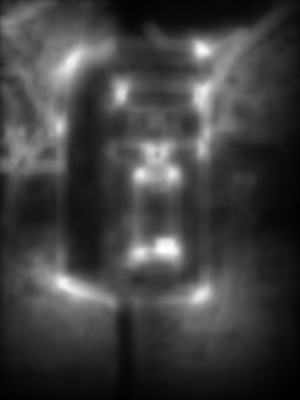} \\

\includegraphics[width = 1.1 cm]{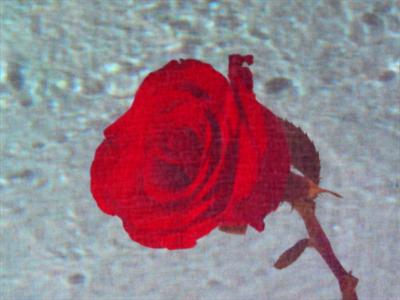}&
\includegraphics[width = 1.1 cm]{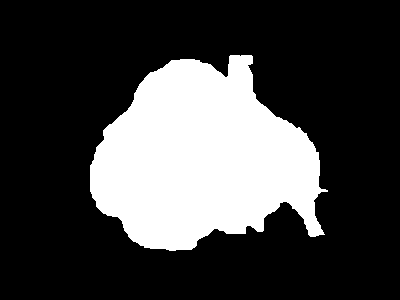}&
\includegraphics[width = 1.1 cm]{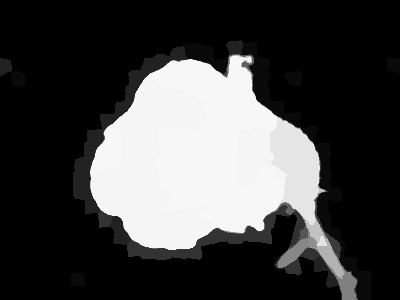}&
\includegraphics[width = 1.1 cm]{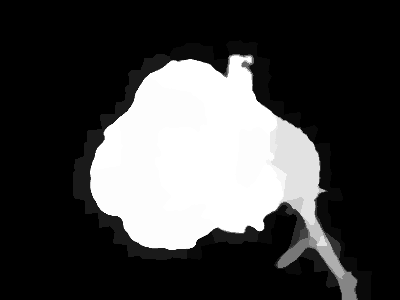}  &
\includegraphics[width = 1.1 cm]{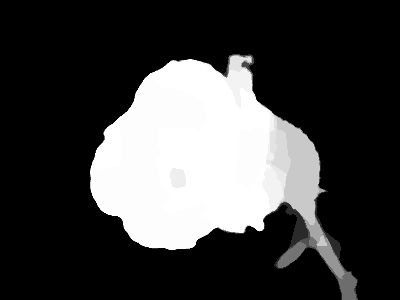}  &
\includegraphics[width = 1.1 cm]{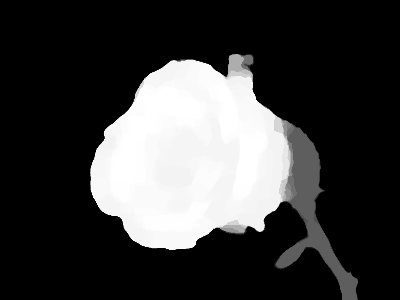}&
   \includegraphics[width = 1.1 cm]{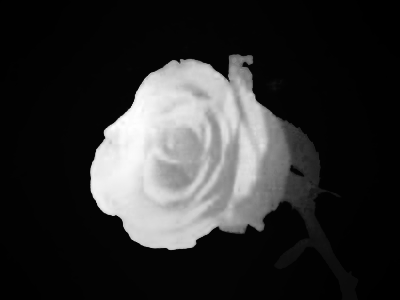}&
   \includegraphics[width = 1.1 cm]{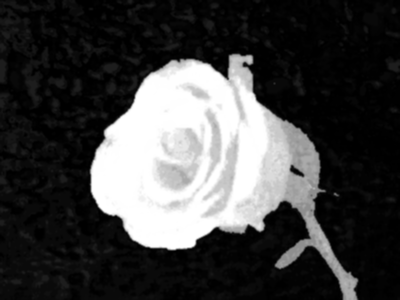} &
   \includegraphics[width = 1.1 cm]{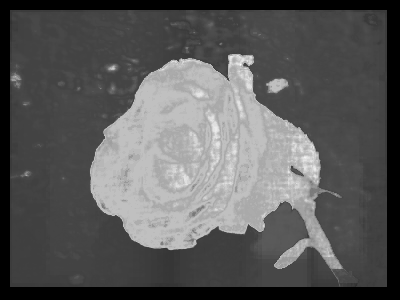} &
      \includegraphics[width = 1.1 cm]{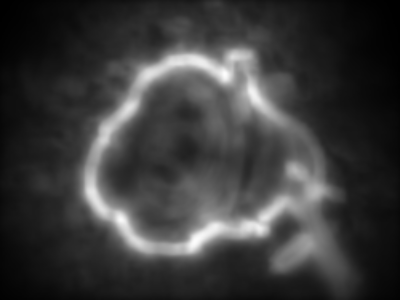} \\

\includegraphics[width = 1.1 cm]{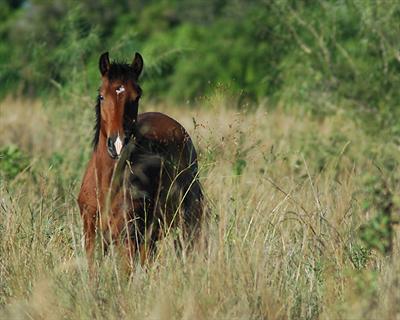}&
\includegraphics[width = 1.1 cm]{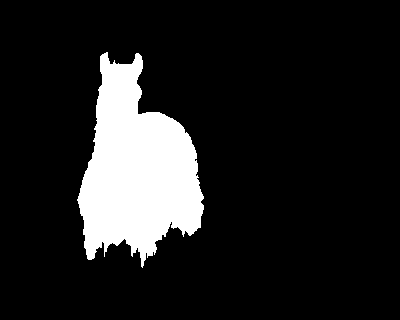}&
\includegraphics[width = 1.1 cm]{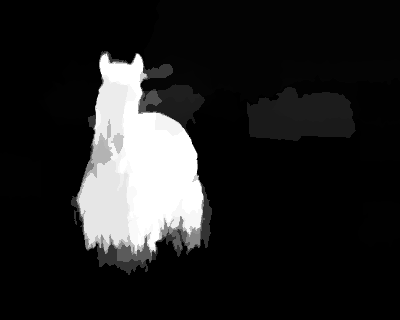}&
\includegraphics[width = 1.1 cm]{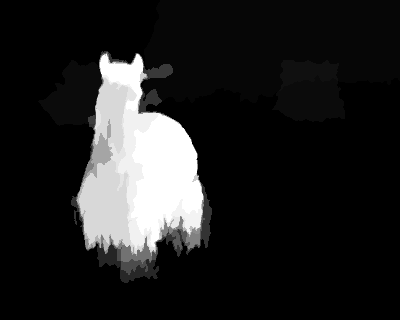}  &
\includegraphics[width = 1.1 cm]{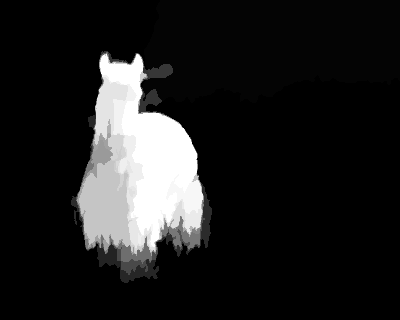}  &
\includegraphics[width = 1.1 cm]{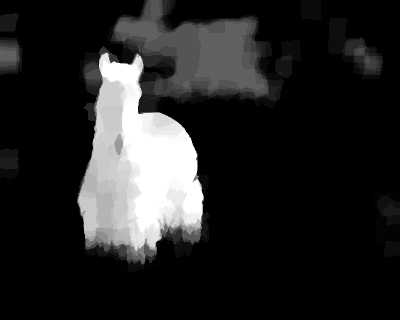}&
   \includegraphics[width = 1.1 cm]{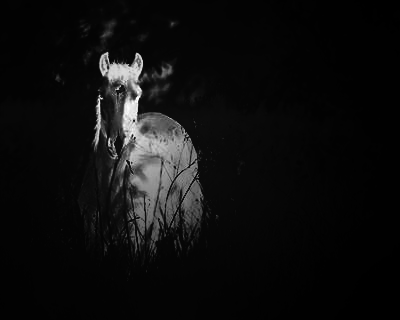}&
   \includegraphics[width = 1.1 cm]{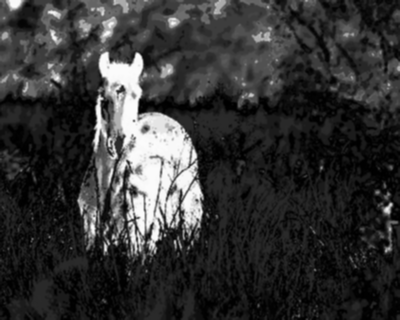} &
   \includegraphics[width = 1.1 cm]{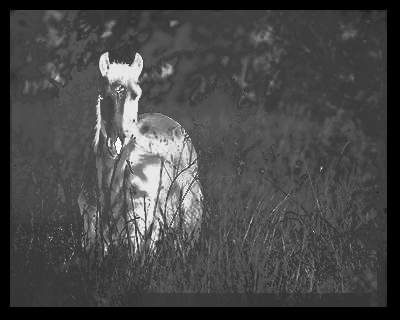} &
      \includegraphics[width = 1.1 cm]{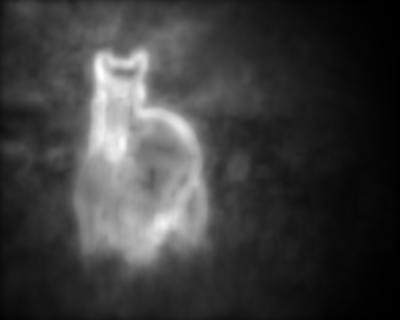} \\

\includegraphics[width = 1.1 cm]{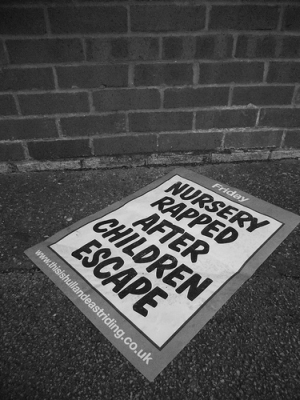}&
\includegraphics[width = 1.1 cm]{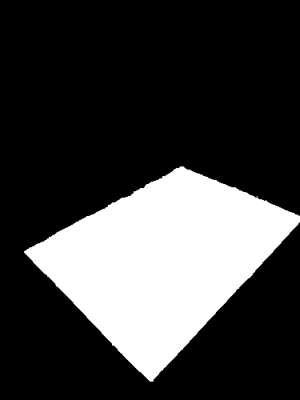}&
\includegraphics[width = 1.1 cm]{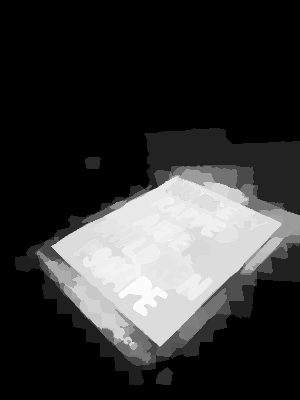}&
\includegraphics[width = 1.1 cm]{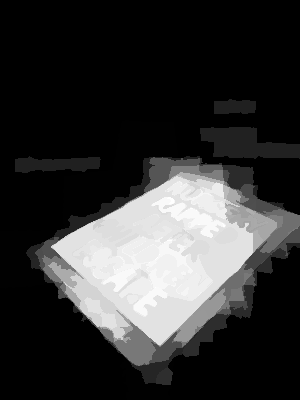}  &
\includegraphics[width = 1.1 cm]{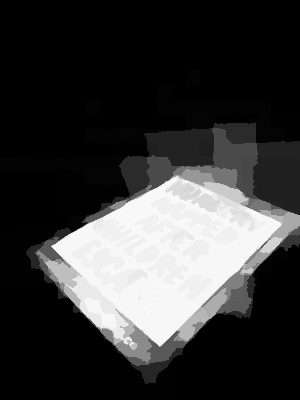}  &
\includegraphics[width = 1.1 cm]{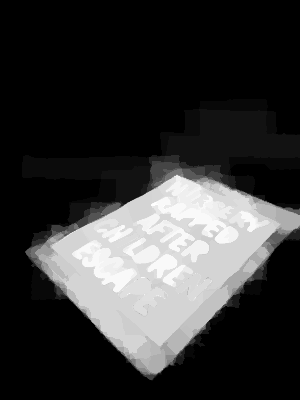}&
   \includegraphics[width = 1.1 cm]{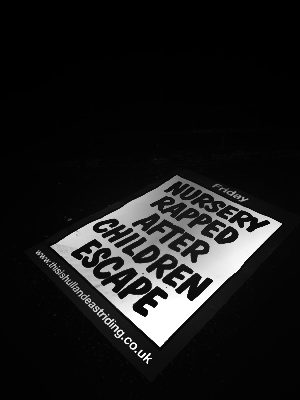}&
   \includegraphics[width = 1.1 cm]{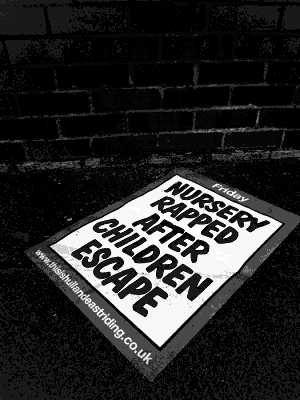} &
   \includegraphics[width = 1.1 cm]{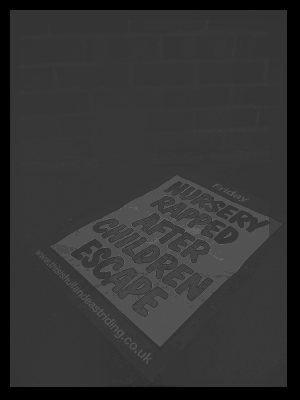} &
      \includegraphics[width = 1.1 cm]{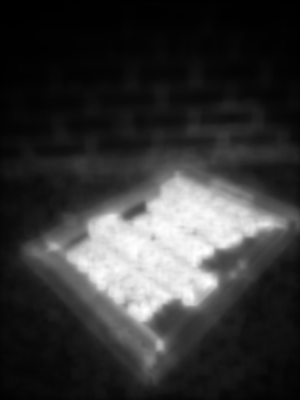} \\

\includegraphics[width = 1.1 cm]{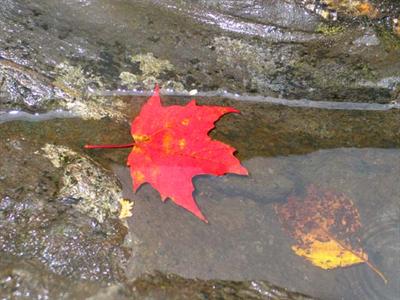}&
\includegraphics[width = 1.1 cm]{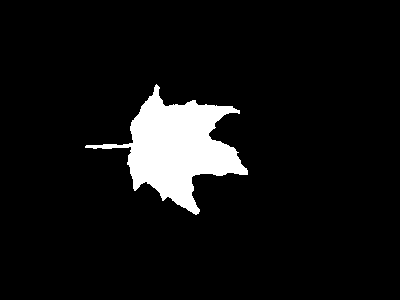}&
\includegraphics[width = 1.1 cm]{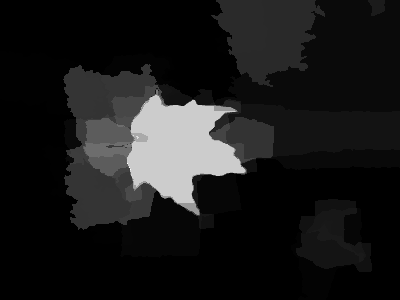}&
\includegraphics[width = 1.1 cm]{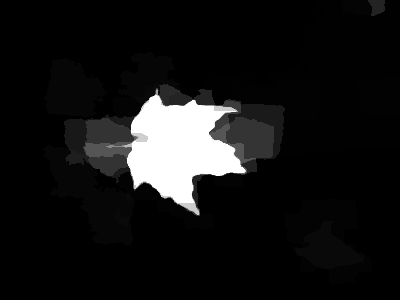}  &
\includegraphics[width = 1.1 cm]{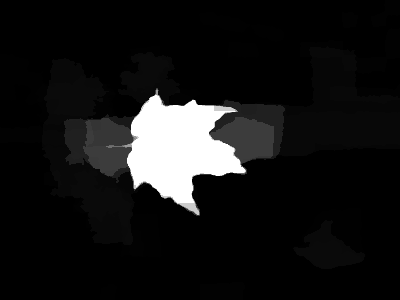}  &
\includegraphics[width = 1.1 cm]{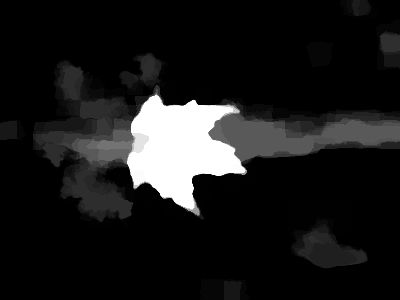}&
   \includegraphics[width = 1.1 cm]{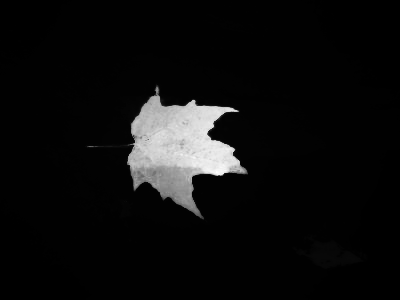}&
   \includegraphics[width = 1.1 cm]{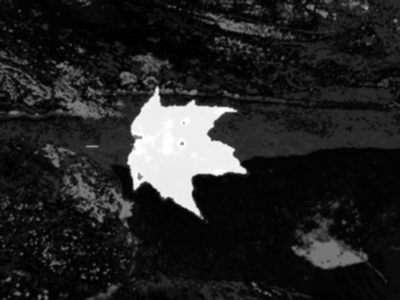} &
   \includegraphics[width = 1.1 cm]{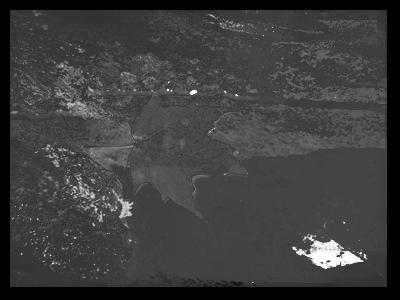} &
      \includegraphics[width = 1.1 cm]{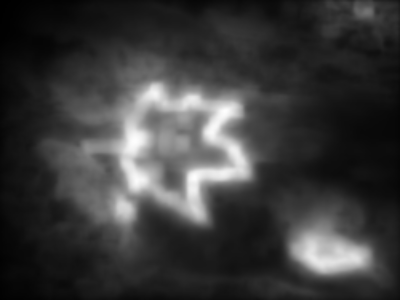} \\
\footnotesize Image &\footnotesize GT& \textbf{\footnotesize C-NeRD}&\textbf{\footnotesize NeRD(75)}&\textbf{\footnotesize NeRD(25)}&\footnotesize SGTD~\cite{scharfenberger2015structure}&\footnotesize \footnotesize SF\cite{perazzi2012saliency}&\footnotesize HC~\cite{cheng2015global}&\footnotesize SM~\cite{SM2010segmenting}&\footnotesize CA~\cite{goferman2012context}
\end{tabular}}
\vspace{-0.2 cm}
\caption{Visual comparison of the proposed NeRD with state-of-the-art salience detection approaches. For these images, NeRD
consistently exceeds saliency performance when visually compared to other unsupervised saliency detection algorithms. }
\label{fig:Res}
\vspace{-0.5 cm}
\end{figure*}

       The relations between atoms are represented by  a statistical model to evaluate their pair-wise neural response divergence given the input image,
with the  pair-wise response of   $\hat{t}_i$ and $\hat{t}_j$  is modeled by $P(\hat{t}_i|\hat{t}_j)$. This conditional probability encodes how probable it is to reconstruct atom $\hat{t}_i$ given atom $\hat{t}_j$. $P(\hat{t}_i|\hat{t}_j)$ can be utilized to evaluate the pair-wise neural response divergence of sparse nodes $i$ regarding node $j$:
\begin{align}
\beta_{ij} =& 1 - P(\hat{t}_i|\hat{t}_j)\\
P(\hat{t}_i|\hat{t}_j) =& \exp \Big(\frac{|\hat{t}_i - \hat{t}_j|}{\sigma^2}\Big) \nonumber
 \end{align}
 where  $P(\cdot)$  formulates the statistical relation  a normal distribution with control parameter $\sigma$.
 \vspace{-0.3 cm}
\subsubsection{Salience Assessment }
 \vspace{-0.2 cm}



The saliency value of region $s_i$ in image $I$ is formulated as the weighted combination of all pairwise salience values $\beta_{ij}$ based on the size of neighbor region $s_j$. A hierarchical approach is taken in to account to preserve the object boundaries more accurately and to result a better saliency map:
\vspace{-0.3 cm}
\begin{align}
\alpha_i  =\sum_{\mathcal{E}_i \in \bar{\mathcal{E}}}  \sum_{j \in \hat{\mathcal{T}}, i \neq j } |s_j| \cdot \beta_{ij}
\end{align}
where $|s_j|$ represents the size of sparse atom  $j$. The salience value $\alpha_i$ of region $s_i$ is propagated to all pixels in region $s_i$ which creates the saliency map. Following~\cite{scharfenberger2015structure}, the saliency map corresponding to different numbers of image elements are aggregated in a hierarchical procedure to preserve fine and coarse image structures while computing the salience value for each region. In contrast to~\cite{scharfenberger2015structure}, which uses 11 hierarchical layers of $\{250, \cdots , 500\}$ atoms where the interval between each two layers is 25  and PCA for dimensionality reduction, NeRD performs well without PCA and only 5 layers of $\{5,\cdots , 85\}$ atoms with 20 as interval. The parameters are optimized based on empirical cross validation. As a result NeRD can provide better performance with a much lower computational cost.

\vspace{-0.4 cm}
\section{Experimental Results}
\label{sec:copyright}
\vspace{-0.3 cm}

The performance of the proposed NeRD approach is evaluated based on two well-known datasets, complex scene saliency dataset (CSSD)~\cite{HSD2013hierarchical} and MSRA10k~\cite{cheng2015global} containing 200  and 10000 images respectively.  NeRD  is also compared with several state-of-the-arts unsupervised salience detection methods including SGTD~\cite{scharfenberger2015structure}, SF~\cite{perazzi2012saliency}, HC~\cite{cheng2015global}, SM~\cite{SM2010segmenting},  and CA~\cite{goferman2012context}  to demonstrate the improvement of the proposed method against hand-crafted feature approaches.

\begin{table}
  \vspace{-0.25cm}
\begin{center}
\footnotesize
    \caption{Area under the curve (AUC), running time, and performance improvement. The proposed NeRD algorithm improves the salience computation by more than 64\% while outperforming SGTD~\cite{scharfenberger2015structure} in terms of AUC.}
        \label{tab:AUC}
    \begin{tabular}{l||cccc}
  ~  &  SGTD  	& C-NeRD	& NeRD(75) & NeRD(25)  \\ \hline \hline
  AUC--CSSD&   0.7784&  0.8124& 0.8141&0.8087 \\
  AUC--MSRA10k& 0.8325  &0.8404 &0.8424 &0.8341  \\ \hline
Running Time (s)&	2.8856	   &   1.2512 &   1.1589 &1.0823           \\
    Improvement &  0     &   56.64\%   &59.84\% &62.49\%     \\
     \end{tabular}
    \vspace{-0.75cm}
\end{center}
\vspace{-0.25cm}
\end{table}
As explained in the previous section the synaptic connectivity percentage of the proposed NeRD here is 25\% which we name the method as NeRD(25). It means that just 25\% of synaptic connectivities in the processing block are selected to extract NeRD features.  It is worth to mention that the percentage of synaptic connectivity is user-defined factor and it depends  on the problem and the demanding speedup  of computational efficiency.
To demonstrate the effect of  synaptic connectivity sparsity of NeRD on salience detection accuracy, NeRD(75) (i.e., with 75\% of synaptic connectivities)  and a specific version of the proposed approach is designed such that all synaptic connectivities are incorporated with zero sparsity factor  in the receptive fields which we name it as conventional NeRD (C-NeRD).

 Figure~\ref{fig:PRCurve} shows  precision-recall curve (PR-curve) for both CSSD and MSRA10k datasets. As seen  NeRD(25) outperforms other unsupervised state-of-the art methods significantly in CSSD dataset while it produce comparable results with SGTD\cite{scharfenberger2015structure} in MSRA10k. Reported results also show that NeRD(25) perform as well as C-NeRD while it is computationally more efficient compared to C-NeRD.
  The StochaticNet receptive field was implemented in MatConvNet~\cite{vedaldi15matconvnet} library. Also to have a fair comparison the salience computation of  NeRD was implemented in a same framework as SGTD while it does not need PCA for feature reduction and also the feature extraction part is a sparse convolution operation which is very fast. Due to these facts, NeRD(25) is much faster than SGTD~\cite{scharfenberger2015structure} framework which is the best one in the competing algorithms (see Table~\ref{tab:AUC} for timing results).

Figure~\ref{fig:Res} demonstrates qualitative comparison of the proposed methods and state-of-the-art algorithms. As seen, NeRD assigns higher salience value to salient object while results smoother saliency map compared to other state-of-the-art approaches. The proposed method produces smoother saliency map with closer value to one for salient object compared to the state-of-the-art algorithms.

  Table~\ref{tab:AUC} reports the area under the curve (AUC) and running time of NeRD compared to SGTD~\cite{scharfenberger2015structure} as the state-of-the-are  with the best performance. The reported results illustrate that the proposed approach outperforms other state-of-the-art methods while improves the computational efficiency by more than 62\%.
\vspace{-0.65 cm}
\section{conclusions}
\vspace{-0.35 cm}
Inspired by the human brain which takes advantage of cues learned from other complex tasks such as object recognition to detect salient object in images, we proposed neural response divergence (NeRD) to extract distinctive feature to do the salience computation. Motivated by StochasticNet framework the sparse representation of synaptic connectivity were provided to address the computational complexity. Results illustrate that NeRD produces more accurate saliency maps while maintains the computational efficiency compared to state-of-the-art methods.

\bibliographystyle{IEEEbib}
\bibliography{refs}

\end{document}